\documentclass[11pt,twocolumn,twoside]{article}
\usepackage[a4paper,top=2cm,bottom=1.5cm,left=1.0cm,right=1.0cm]{geometry}
\usepackage[english]{babel}
\usepackage[T1]{fontenc}
\usepackage{lmodern}
\usepackage[utf8]{inputenc}
\usepackage[scaled=0.95]{helvet}
\usepackage{mathptmx}
\usepackage{amsmath}
\usepackage{graphicx}
\usepackage{xcolor,calc}
\usepackage[colorlinks]{hyperref}
\hypersetup{linkcolor = black, citecolor = black, urlcolor = blue} 
\usepackage{microtype}
\usepackage{sectsty}
\usepackage[font={small},labelfont={sf,bf}, margin=0cm, indention = 0cm]{caption}
\usepackage{titling}
\usepackage{authblk}
\usepackage{booktabs}
\usepackage{fancyhdr}
\usepackage{setspace}
\usepackage[minnames=3,maxnames=3,isbn=false,url=false,eprint=false,giveninits=true,sorting=none,citestyle=numeric-comp,backend=bibtex]{biblatex}
\usepackage{csquotes}
\usepackage{amssymb}

\pretitle{\begin{center}\sffamily\LARGE\bfseries}
\posttitle{\par\end{center}} 

\predate{}
\postdate{}
\date{}

\allsectionsfont{\sffamily\bfseries\normalsize}

\newenvironment{customabstract}
    {\footnotesize 
     \begin{tabular}{p{0.95\columnwidth}} 
     \textbf{Abstract}
    }
    {
    \end{tabular} 
    \vspace{-0.3cm}
    }

\renewbibmacro{in:}{}    

\AtBeginBibliography{\footnotesize}

\begin{document}

\title{ 
Low Dose Helical CBCT denoising by using domain filtering with deep reinforcement learning } 

\author[1]{Wooram Kang}
\author[1]{Mayank Patwari}

\affil[1]{Department of Medical Engineering,
          Friedrich Alexander Universität Erlangen-Nürnberg, Germany}

\maketitle
\thispagestyle{fancy}


\begin{customabstract}
Cone Beam Computed Tomography(CBCT) is a now known method to conduct CT imaging. Especially, The Low Dose CT imaging is one of possible options to protect organs of patients when conducting CT imaging. Therefore Low Dose CT imaging can be an alternative instead of Standard dose CT imaging. However Low Dose CT imaging has a fundamental issue with noises within results compared to Standard Dose CT imaging. Currently, there are lots of attempts to erase the noises. Most of methods with artificial intelligence have many parameters and unexplained layers or a kind of black-box methods. Therefore, our research has purposes related to these issues. Our approach has less parameters than usual methods by having Iterative learn-able bilateral filtering approach with Deep reinforcement learning. And we applied The Iterative learn-able filtering approach with deep reinforcement learning to sinograms and reconstructed volume domains. The method and the results of the method can be much more explainable than The other black box AI approaches. And we applied the method to Helical Cone Beam Computed Tomography(CBCT), which is the recent CBCT trend. We tested this method with on 2 abdominal scans(L004, L014) from Mayo Clinic TCIA dataset We observed. The results and the performances of our approach overtake the results of the other previous methods.

\end{customabstract}

\section{Introduction}

There are various ways to conduct computed tomography. and Helical Cone Beam Computed Tomography(CBCT) is one of recent methods to conduct computed tomography imaging. however The Reconstruction of Helical CT from the native geometry is a difficult problem. Therefore, there are several rebinning method from native cone beam geometry to Fan beam geometry, Parallel beam geometry or the other known geometry to translate the difficult problem to a well-known problem. \cite{1}\cite{2}\cite{3}\cite{4}

On the other hand, the reconstruction of helical cone beam computed tomography sinogram has some an additional issue to consider, which is Flying Focal Spot(FFS) problem. In the reconstruction process, the focal spot from the curved detector to source of the CT machine can have a difference between the ideal focal spot and the experimental focal spot. By having the difference, the reconstruction result can be noisy because of the mismatch of focal spot. multiple sampling and approximation from nearest projections based on the direction of Flying Focal Spot(FFS) deflection can be an solution to handle Flying Focal Spot(FFS) Deflection. \cite{5}\cite{7}

Noise removal in CT imaging is usually performed in conjunction with the reconstruction process. Most of Iterative reconstruction algorithms including Siemens ADMIRE and Canon AIDR. \cite{8}\cite{9} 

Deep learning approaches have been successfully applied to the CT denoising problem. Most of deep learning approaches for CT denoising are formulated in the form of image translation tasks \cite{10} \cite{11} \cite{12} \cite{13} \cite{14} 

But there is a fundamental problem with deep learning approach with image translation or other image modal based methods, which is lack of medical data to train to solve specific problems and heavy deep learning model with a large amount of parameters. 

Therefore, An approach including less parameters than previous approaches and having meaningful results even with small amount of medical data is necessary. In this paper, we would discuss about the a deep reinforcement learning with Iterative bilateral filtering. And we applied this method to the target sinograms and the volume domains. The approach has a notably small amount of inference parameters and it can be applied with a small amount of Cone Beam Computed Tomography(CBCT) data. The performance is also quite reasonable. It can be proved with the results and the metrics followed in the results section.

\section{Materials and Methods}

\begin{figure*}
  \centering
  \includegraphics[width=1.0\textwidth, height=0.20\textheight]{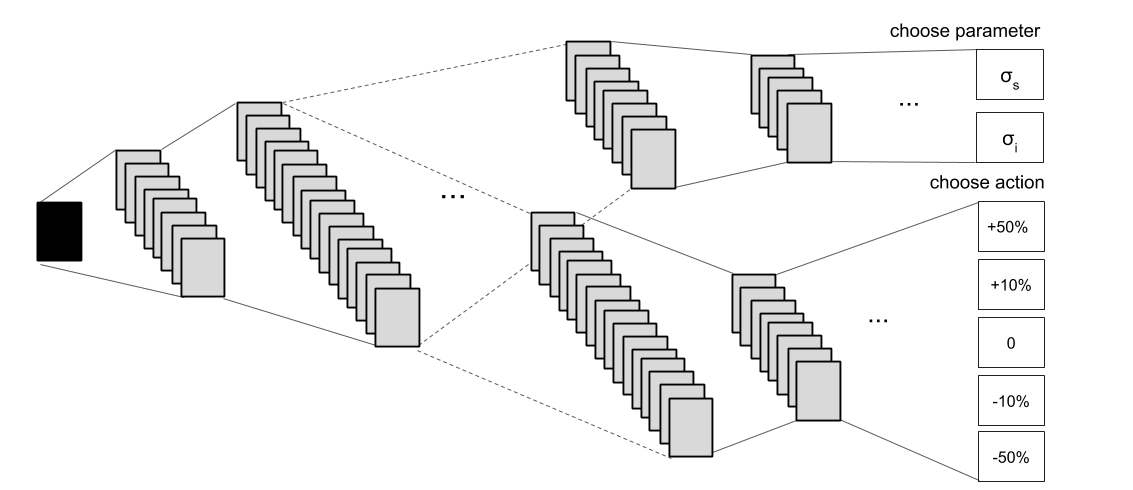}
  \caption{General Neural Network Architecture of Bilateral Filtering with Deep Q learning}
  \label{fig:brief_method_2}
\end{figure*}

\subsection{Helical CBCT reconstruction}

Based on the cone beam native geometry, the 2D view weighted CB-FBP reconstruction algorithm is expressed as

\begin{equation}
\hat{f_{n}}(x,y,z) = \dfrac\pi{({\beta_{\max}} - {\beta_{\min}})} \int_{\beta_{\min}} ^ {\beta_{\max}} \dfrac{R^2}{L^2(x,y,\beta)}w(\beta,\gamma)\hat{p}(\alpha,\beta,\gamma)d\beta
\end{equation}

\begin{equation}
\hat{p}(\alpha,\beta,\gamma)=\cos{\alpha}[(\cos{\gamma} * p(\alpha,\beta,\gamma)  \circledast g(\gamma)]
\end{equation}

\begin{equation}
L(x,y,\beta)=\sqrt{(R+x\cos{\beta} + y\sin{\beta})^2 + (-x\sin{\beta} +y\sin{\beta})^2}
\end{equation}

Where g is 1D filtering, eg. ramp filtering. p is projection. beta stands for the maximum and the minimum of source trajectory angles. w is weighting function.

And Based on the cone beam parallel geometry after rebinning, the 2D view weighted CB-FBP reconstruction algorithm is expressed as

\begin{equation}
f_{n}(x,y,z) = \dfrac\pi{({\beta_{\max}} - {\beta_{\min}})} \int_{\beta_{\min}} ^ {\beta_{\max}} \dfrac{R^2}{ \sqrt{R^2 + Z^2}}w(\beta,t)\hat{s}(\alpha,\beta,t)d\beta
\end{equation}

\begin{equation}
\hat{s}(\alpha,\beta,t)= s(\alpha,\beta,t)  \circledast g(\gamma)
\end{equation}

Where Z is the projected z-coordinate of point P(x, y, z) onto the virtual detector. 1D filtering specified by equation becomes tangential automatically due to the row-wise fan-to-parallel rebinning. and As the equation after rebinning get simpler, The Reconstruction with rebinning can be performed with less computing cost in the practical experiments. and it can be implemented easily with inverse radon transform with weighting.

But helical Cone Beam Computed Tomograph(CBCT) can have a fundamental issue on projections because of flying focal spot. When the helical Cone Beam Computed Tomograph(CBCT) scanner scan patients, there is possibility to have deflection between the ideal focal spot and the practical focal spot for each projection. so it can make noise to projection and it results in having reconstruction results with noise which looks like light emission. 

To handle the issue, there are multiple sampling and subfan approximation technique we can use. 
\begin{equation}
S \leftarrow s(\alpha) = s(\alpha, \partial\alpha, \partial z)
\end{equation}
\begin{equation}
D\leftarrow r(\alpha, \beta, t) = s(\alpha) + R\begin{pmatrix}{-\sin{(\alpha + \beta)}}\\ \cos{(\alpha + \beta)}\\0\end{pmatrix} + t\begin{pmatrix}0\\0\\1\end{pmatrix}
\end{equation}
R is the distance between source and detector. Spiral source trajectory define as S and Cylindrical detector trajectory define as D. to neutralize the deflection, we can sample projections from 4 different sources and interpolate the readings. 
\begin{equation}
S_n\leftarrow s(\alpha \pm\dfrac{1}{2}\Delta\alpha^{phys}\pm\Delta\alpha^{phys}, \pm\partial\alpha, \pm\partial z)
\end{equation}
Usually, after multiple sampling, there is  an additionally necessary process, which is subfan approximation. but our reconstruction is based on rebinning to parallel geometry we do not need to use the sub-fan approximation. The rebinning process account for the odd geometry of the Flying Focal Spot(FFS) scanner.
\begin{figure*}
  \centering
  \includegraphics[width=0.90\textwidth,height=0.21\textheight]{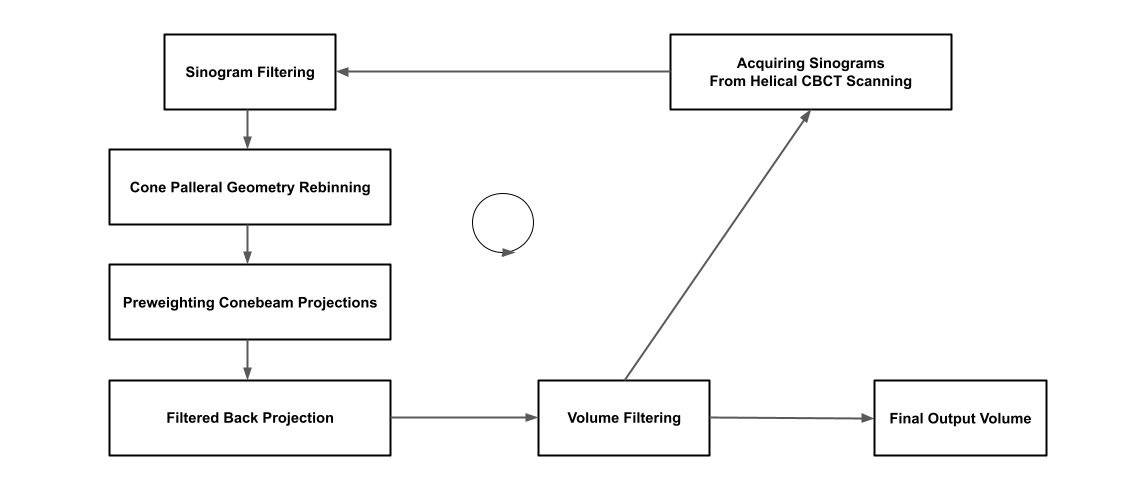}
  \caption{Workflow of Iterative denoising of helical CBCT with bilateral filtering to sinograms and volume domains with deep Q learning}
  \label{fig:brief_method}
\end{figure*}
\subsection{bilateral filtering with Deep Q learning}
Figure \ref{fig:brief_method_2} shows general neural network architecture of Bilateral as described below. Filtering with Deep Q learning. First of all, The bilateral filter can be formulated as 
\begin{equation}
I_f(x)=\displaystyle\frac{\sum_{o\in N(x)}{I_n {G_{\sigma_s}} (x - o) {G_{\sigma_i}} (I_n(x) - I_n(o)) }} { \sum_{o\in N(x)}{ {G_{\sigma_s}} (x - o) {G_{\sigma_i}}(I_n(x) - I_n(o) ) }}
\end{equation}
\begin{equation}
G_{\sigma}(x)=\dfrac{e^{-x^2/2\sigma^2 }}{2 \sigma^2}
\end{equation}

Where sigma is a hand tuned parameter, which determines the strength of filtering. There are two sigma parameters in a bilateral filter, sigma i which controls the strength of the difference of intensities and sigma s which controls the strength of the difference of spatial coordinates. so there would be 2 pairs of sigma i and sigma s for sinogram and volume domains.
The filter for sinogram consider 5 * 5 neighborhoods and the filter for volume domain consider 5 * 5 * 5 neighborhoods in our experiment.

And the deep Q learning network train the sigmas with the architecture in Figure 1. General formular of Policy can be described as 
\begin{equation}
Q^*(s, a) = \max_\pi[r^0 + \alpha r^{1} + ... + \alpha^n r^{n} |s_k =s, a_k=a ,\pi]
\end{equation}
A property of Q(s, a), as described by Bellman equation is the following 
\begin{equation}
Q^*(s, a) = r+ \alpha  \max_a Q^*(s' , a')
\end{equation}
loss function for the bellman equation can be defined as 
\begin{equation}
L(W) = [r + \alpha  Q \max_a (s', a' ; W) - Q(s, a ;W)] ^2 
\end{equation}
for the modern deep Q learning, There is double deep Q learning. so finally we use the loss function of the double deep Q learning described as below.
\begin{equation}
L(W) = [r + \alpha Q(s',  Q \max_a (s', a' ; W); W') - Q(s, a ;W)] ^2 
\end{equation}
and For the target reward network, we defined the target reward function as
\begin{equation}
T(I) = GSSIM(I, I_{gt}) + \dfrac{1}{ \dfrac{Average((I - I_{gt})^2) }{ROI} + 1 }    
\end{equation}
and the reward network consists of small simple CNN with 3 *3 kernels. 
The Policy Networks are made with simple CNN with 3 * 3 kernel and 128 neurons fully connected layer for sinogram denoiser and simple CNN with 3 * 3 * 3 kernel and 128 neurons fully connected layer for Volume denoiser.

\subsection{Iterative Denoising with Learnable bilateral Filtering}
First of all, the process starts with scanning obviously. after that, the sinograms acquired from scanning has to be filtered. and then, the sinograms of the helical Cone Beam Computed Tomograph(CBCT) will be rebinned to Cone parallel Geometry from Cone beam native geometry. The rebinned sinograms will be preweighted and Filtered Back Projection will be conducted with the sinograms. Therefore, we can have a Volume from filtered sinograms. then, we can apply image or volume domain filtering to the volume. Therefore, We have a volume filtered twice with sinogram domain and volume domain. If the results don't look good enough, then Forward Projection will be conducted and the sinogram from the forward projection will be given to the input of sinogram filtering again. To have Final output volume of the process, we can conduct the cycle iteratively from sinogram filtering to Forward Projection til finishing the cycle the selected iteration number of time or having good enough results. Figure \ref{fig:brief_method} shows The practical workflow overview of our research as described.

\subsection{Computing environment and runtime measuring}
the experiment machine has Intel i-7 9850H, 16GB RAM, 4GB GPU RAM Quardro T2000. With this setup, sinogram reading took 20 seconds, rebinning took 4 seconds, preweighting took 4 seconds, Filtered Back Projection took 171 seconds, sinogram filtering took 10 seconds, Volume filtering took 54 seconds. the parts From sinogram reading to filtered back projection are implemented by CPU-oriented and the spent times are calculated based on each rotation of scanning. and by using Jit framework, the calculations of CPU-oriented parts were paralleled to some points. 

\section{Results}

\begin{figure}
  \centering
  \includegraphics[width=\columnwidth]{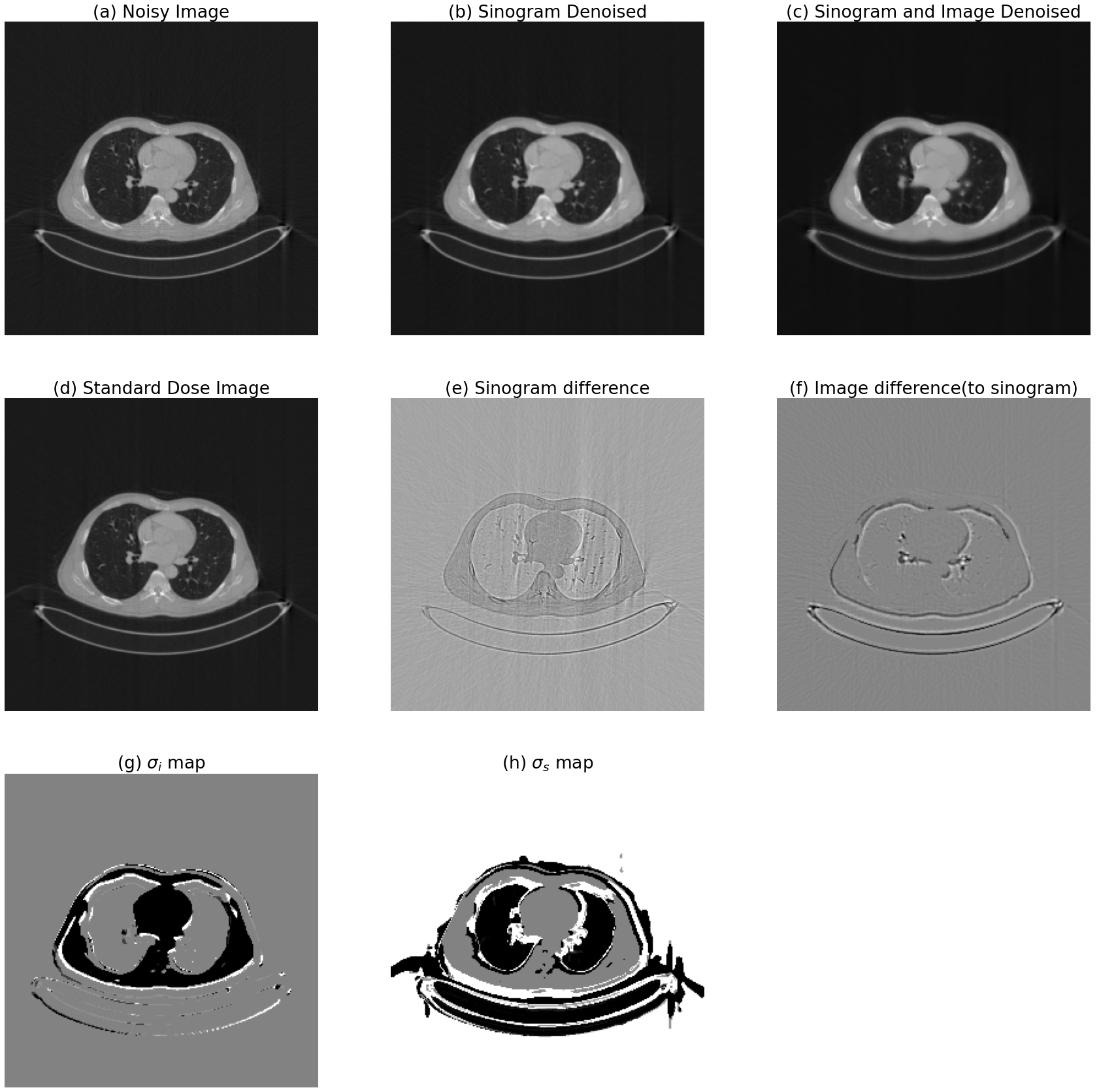}
  \caption{Denoising results compared to Standard dose and details}
  \label{fig:result_method}
\end{figure}

\begin{figure}
  \centering
  \includegraphics[width=\columnwidth]{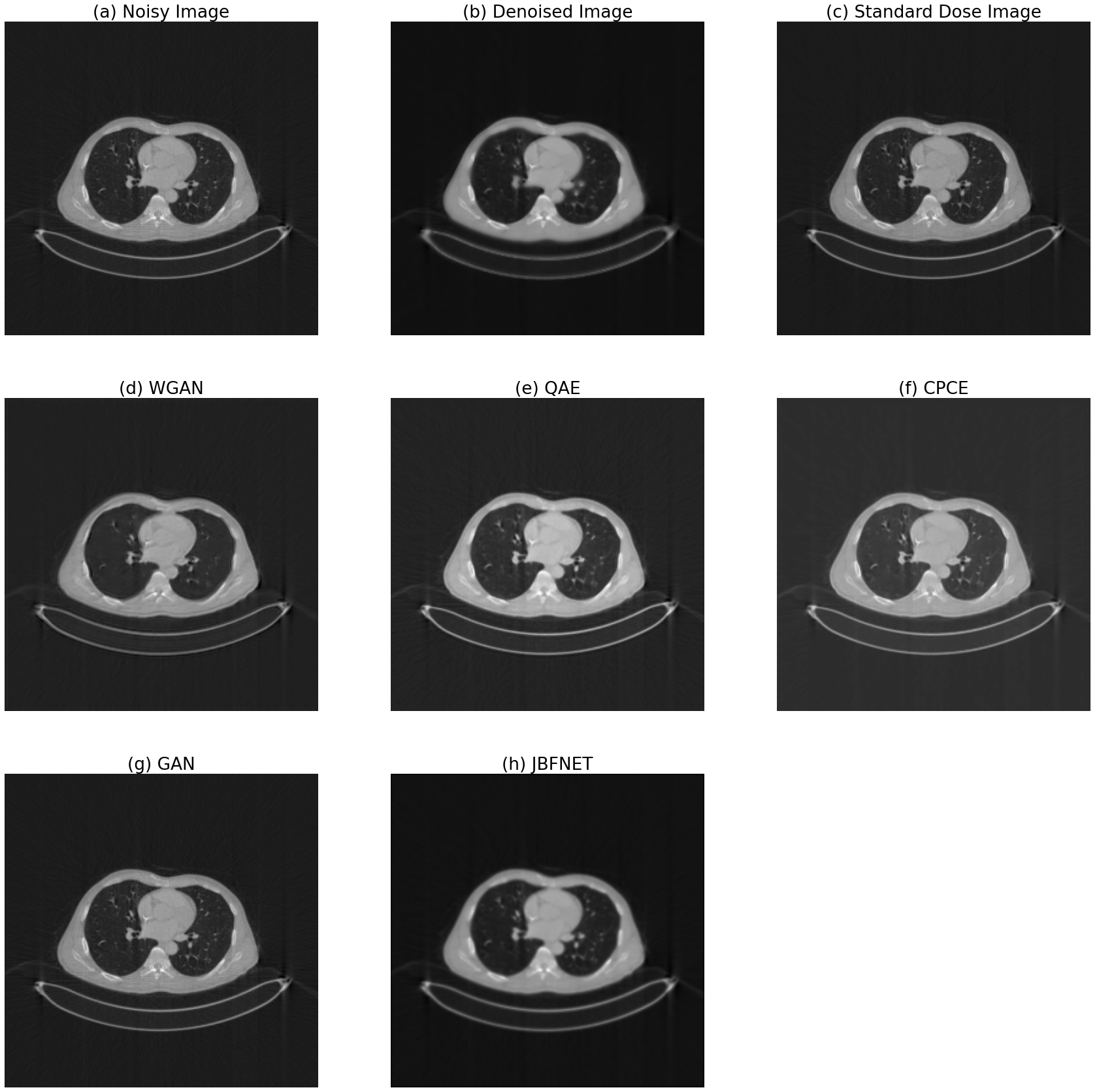}
  \caption{Denoising results compared to other methods}
  \label{fig:result_compare}
\end{figure}

\subsection{Metrics and Evaluation}
Figure \ref{fig:result_method} shows that the differences between the standard dose image and the denoised image and sigma map for filtering. We can know where and how much the image has been filtered by our approach. %

Figure \ref{fig:result_compare} shows the results from our approach and other methods pretrained with other CT images. the results can be compared to others.

We can find that JBFNet and this approach were robust to some points. It is because the others are basically based on image translation neural network models. However our approach and JBFNet learn parameters for filtering. therefore, it doesn't get effected so much on the difference of geometries or image range, etc. Therefore, those approaches can have comparably robust results.

There is a proof for the image domain range issue in Figure \ref{fig:result_compare}. after matching the image domain range to how much the models are pre-trained with training dataset with. the images quality got better notably. it can be one of future benefits of using this approach. this method can be robust to any range of HUs so that it can be applied to general problems. before matching the domain, the metric values including PSNR and SSIM of the GAN-based models don't look good, which means GAN-based models are over-sensitive to the image domain range and geometries of training data. and on the other hand, QAE, JBFNet, this approach are robust from image domain range issue. 

\begin{figure}
  \centering
  \includegraphics[width=\columnwidth]{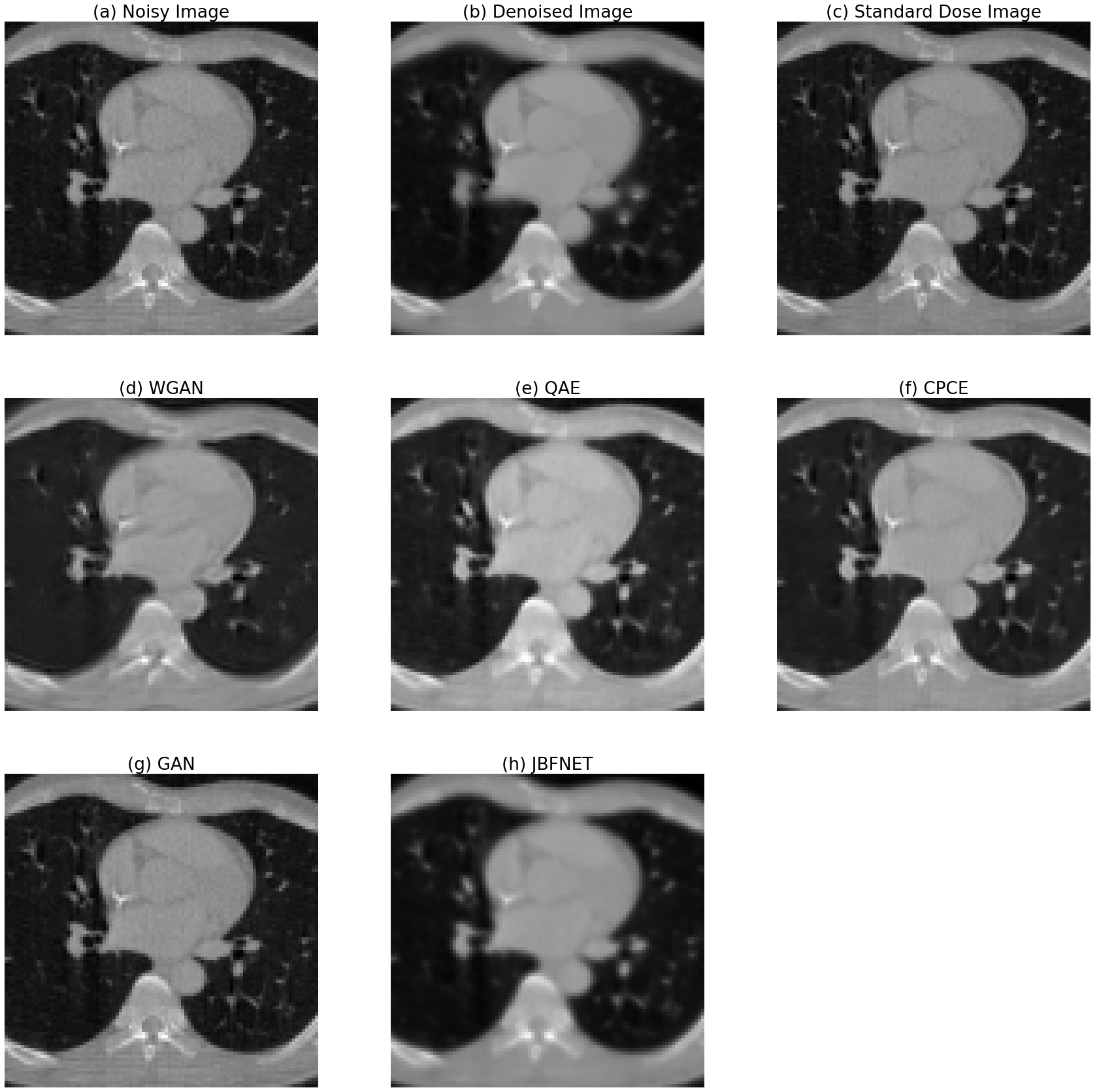}
  \caption{Specific areas of results including results from other methods}
  \label{fig:result_compare_after_zoom}
\end{figure}

In Figure \ref{fig:result_compare_after_zoom}, we can compare specific areas of results. There is a disadvantage of models based on domain filtering including our works and JBFNET. the image in the middle is clearer than others but the outside of images are a bit blurry. 

\begin{figure}
  \centering
  \includegraphics[width=1.00\columnwidth]{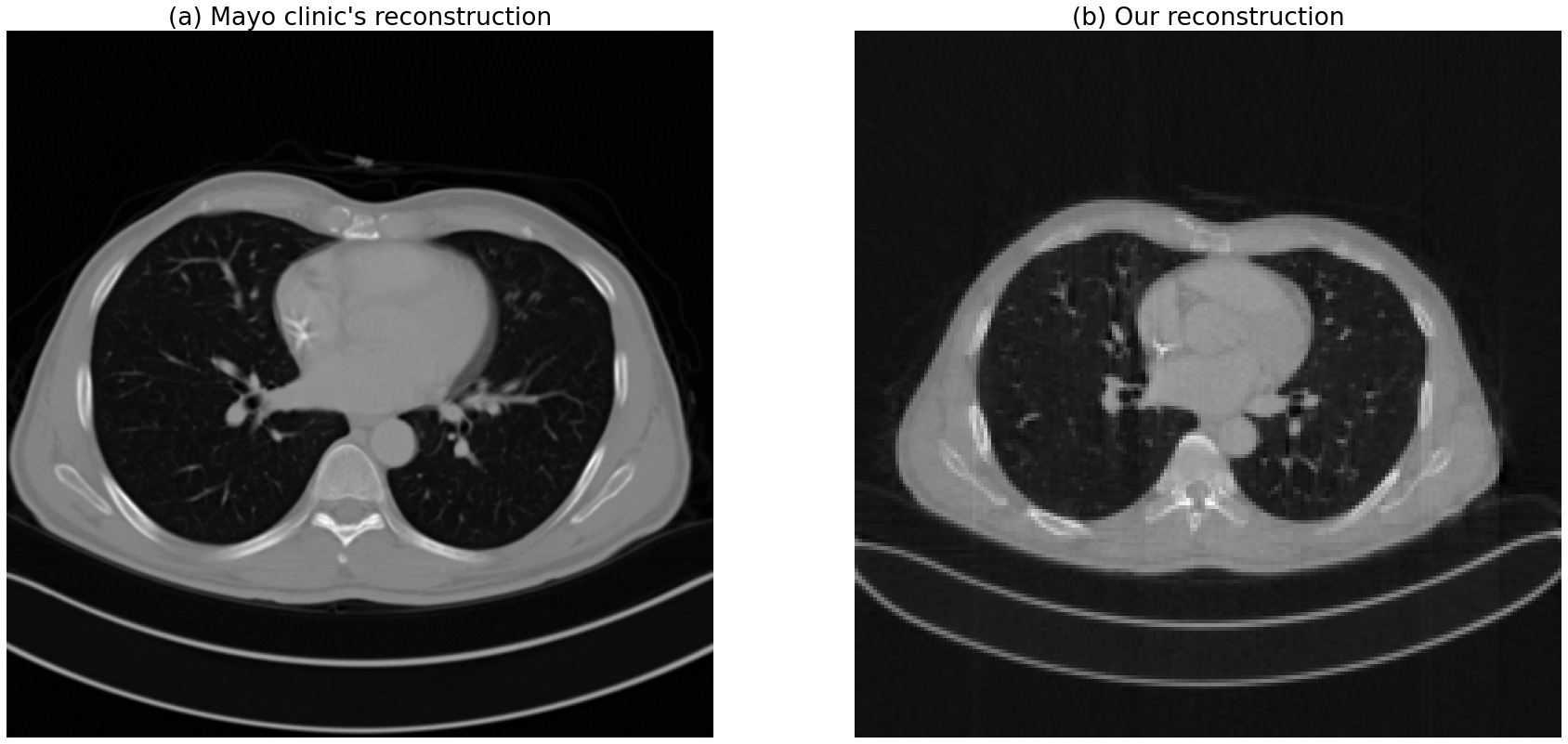}
  \caption{Mayo clinc standard dose reconstruction data sample}
  \label{fig:gt__shot}
\end{figure}

And the numbers of parameters to train are also notably different for each model. JBFNET and our approach learn the proper parameters for filtering so the numbers of parameters to inference can be really small, which are parameters only for filtering. however the other methods need still big models with the large numbers of parameters to inference. Table \ref{tab:paramsTable} shows the numbers details.

the quality of the results was hard to be compared directly to mayo clinic reconstruction results. because as we can see in Figure \ref{fig:gt__shot}, extra post-processing methods were applied to the volume obviously. and the reconstruction range is narrow than our results so it is also another reason to make it harder.

In the results quality point of view, Table \ref{tab:COMPAREdummytable} shows that each PSNR and SSIM of the results compared to the standard dose. and the values are means of overall experiments. overall qualities was also interesting. it is same or similar to low dose Cone Beam Computed Tomograph(CBCT) results which means this approach can preserve features for the patients well and also the denoising quality is also good. 

\begin{table}
  \centering
  \begin{tabular}{lrrr}
    \toprule
    \textbf{Method}      & \textbf{PSNR}      & \textbf{SSIM}   \\
    \midrule
    LOW DOSE CBCT          & 41.79       & \textbf{0.96368}        \\
    
    \midrule
    
    WGAN\cite{8340157}          & 32.71       & 0.93477      \\
    QAE\cite{13}          & \textbf{41.85}       & 0.97584        \\
    GAN\cite{10}          & 39.89       & 0.96322        \\
    CPCE\cite{11}          & 38.60       & \textbf{0.98473}       \\
    JBFNET\cite{14}          & 36.78       & 0.96879     \\
    
    \midrule
    
    THIS METHOD*         & \textbf{35.38}       & \textbf{0.96251}  \\  
  \end{tabular}
  \caption{PSNR and SSIM results **compared to standard dose}
  \label{tab:COMPAREdummytable}
\end{table}

\begin{table}
  \centering
  \begin{tabular}{lrrr}
    \toprule
    \textbf{Method}      & \textbf{In Training}      & \textbf{Inference}   \\
    \midrule
    WGAN\cite{8340157}          & 1,464,546       & 56,097        \\
    QAE\cite{13}          & 137,658       & 137,658        \\
    GAN\cite{10}          & 1,375,386       & 861,857        \\
    CPCE\cite{11}          & 118,209       & 118,209       \\
    JBFNET\cite{14}          & 118,697       & 488     \\
    \midrule
    \textbf{THIS METHOD*}         & \textbf{1,002,946}       & \textbf{4}  \\  
  \end{tabular}
  \caption{Numbers of parameters for each model}
  \label{tab:paramsTable}
\end{table}

\section{Discussion And Conclusion}

In this paper, this paper showed the helical Cone Beam Computed Tomograph(CBCT) reconstruction and The Helical Cone Beam Computed Tomograph(CBCT) reconstruction can have potential issues including matching geometries, flying focal spot, etc. And we showed that rebinning to cone parallel geometry makes the reconstruction easier and also handling flying focal spot easier. 

Additionally, for handling flying focal spot, we showed multiple sampling methods to estimate ideal focal spots. It can be simpler with having rebinning to cone parallel geometry. Because we do not need subfan approximation with the parallel geometry rebinning.

Deep reinforcement learning was also applied for denoising. And domain filtering with the deep reinforcement learning can be useful to have less number of parameters and to make meaningful results with the small model. And compared to GAN-based models, approaches with filtering were much robust regardless of the effects of training data. 

There can be the future works with those results. The filtered back projection calculation time in our approach can be improved as well. In our experiment, it is CPU-oriented implementation so it can be improved by having GPU implementation. The number of parameters for testing is really small but the number of parameters for training is still large. So to make smaller numbers of parameters like other methods. And the quality of images were good but there are blur areas coming from bilateral filtering. The approach can be improved to avoid making those areas.

\newpage
\printbibliography

\end{document}